\documentclass[10pt,journal,compsoc]{IEEEtran}
\ifCLASSOPTIONcompsoc
  \usepackage[nocompress]{cite}
\else
  \usepackage{cite}
\fi
\ifCLASSINFOpdf
\else
\fi
\usepackage{algorithm}
\usepackage{algorithmic}
\usepackage[switch]{lineno} 
\usepackage{epsfig}
\usepackage{amsmath}
\usepackage{amssymb}
\usepackage{color}
\usepackage{bm}
\usepackage{multirow}
\usepackage{booktabs}

\begin{document}
%
\title{Image-to-Character-to-Word Transformers for Accurate Scene Text Recognition}

\author{Chuhui Xue, Jiaxing Huang, Wenqing Zhang, Shijian Lu$^*$, Changhu Wang, Song Bai
\IEEEcompsocitemizethanks{\IEEEcompsocthanksitem Chuhui Xue, Jiaxing Huang and Shijian Lu are with the School of Computer Science and Engineering, Nanyang Technological University, Singapore.\protect\\
\IEEEcompsocthanksitem Wenqing Zhang, Changhu Wang and Song Bai are with Bytedance Inc.\protect\\
\IEEEcompsocthanksitem E-mail: xuec0003@e.ntu.edu.sg, jiaxing.huang@ntu.edu.sg, zhangwenqin\\g.gordon@bytedance.com, shijian.lu@ntu.edu.sg, changhu.wang@gmail.c\\om, songbai.site@gmail.com.\protect\\
\IEEEcompsocthanksitem $*$ denotes corresponding author.\protect\\
\IEEEcompsocthanksitem Work done during an internship at ByteDance Inc.
}}

\IEEEtitleabstractindextext{%
\begin{abstract}
Leveraging the advances of natural language processing, most recent scene text recognizers adopt an encoder-decoder architecture where text images are first converted to representative features and then a sequence of characters via `sequential decoding'. However, scene text images suffer from rich noises of different sources such as complex background and geometric distortions which often confuse the decoder and lead to incorrect alignment of visual features at \textit{noisy decoding time steps}. This paper presents I2C2W, a novel scene text recognition technique that is tolerant to geometric and photometric degradation by decomposing scene text recognition into two inter-connected tasks. The first task focuses on image-to-character (I2C) mapping which detects a set of character candidates from images based on different alignments of visual features \textit{in an non-sequential way}. The second task tackles character-to-word (C2W) mapping which recognizes scene text by decoding words from the detected character candidates. The direct learning from character semantics (instead of noisy image features) corrects falsely detected character candidates effectively which improves the final text recognition accuracy greatly. Extensive experiments over nine public datasets show that the proposed I2C2W outperforms the state-of-the-art by large margins for challenging scene text datasets with various curvature and perspective distortions. It also achieves very competitive recognition performance over multiple normal scene text datasets.
\end{abstract}

}

\maketitle

\IEEEdisplaynontitleabstractindextext

%
\IEEEpeerreviewmaketitle

\section{Introduction}\label{sec:introduction}

\begin{figure}[t]
  \centering
  \includegraphics[width=\linewidth]{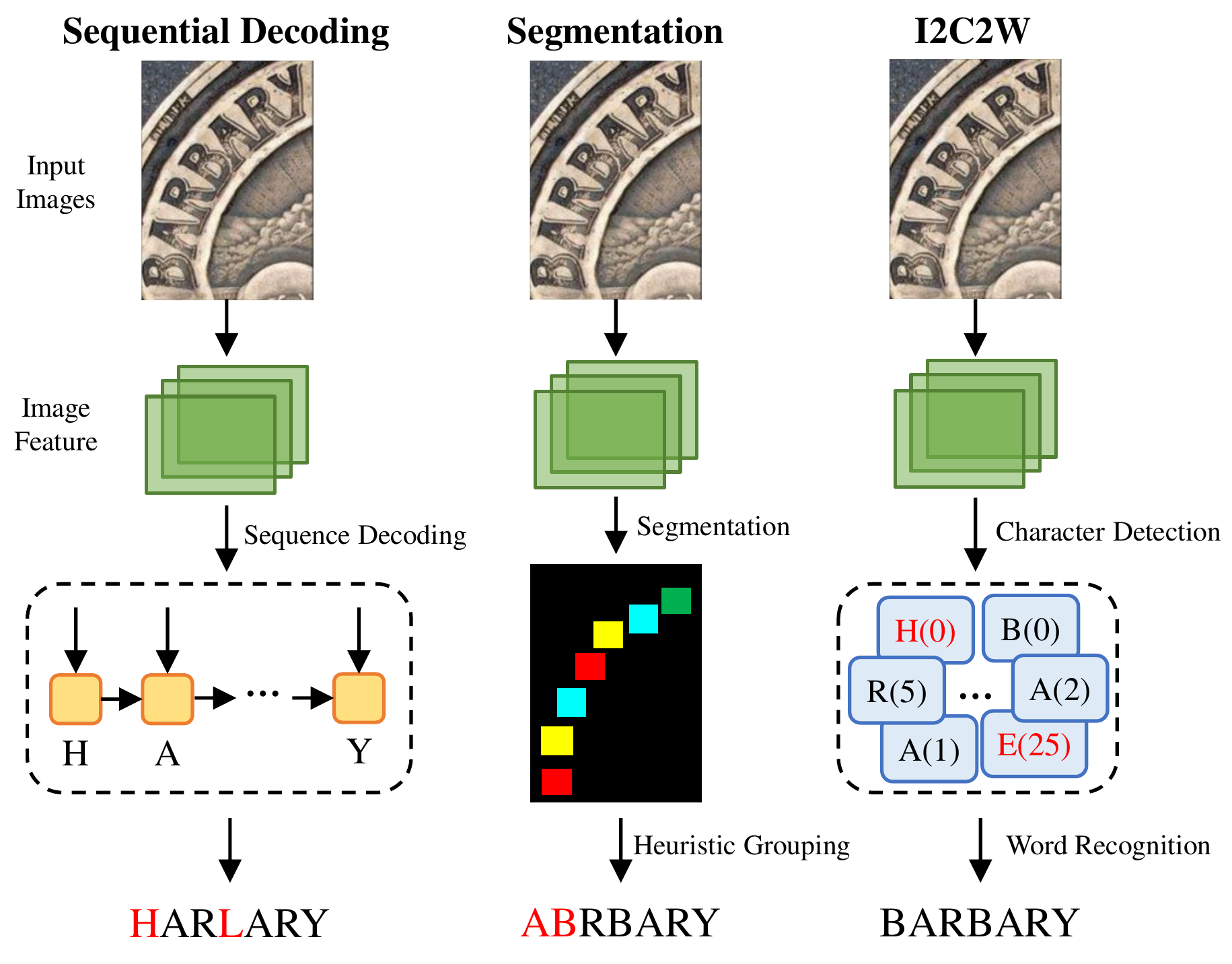}
\caption{
\textbf{Illustration of different scene text recognition pipelines}: \textit{Sequential Decoding} approach decodes a sequence of characters directly from visual features which is prone to misalignment and incorrect predictions due to geometric distortions or noisy background. \textit{Segmentation} approach requires character-level annotations in training and it tends to fail due to its adopted heuristic character grouping rules. Differently, the proposed \textit{I2C2W} first detects character candidates as illustrated in blue boxes (including false detection as highlighted in red color) and predicts their relative positions (indicated by digits within the bracket following the recognized characters). It then recognizes words from the character candidates (instead of visual features) by learning character semantics and correcting falsely detected characters.
}
\label{fig:intro}
\end{figure}

Texts in scenes contain rich semantic information that is very important and valuable in many practical applications such as autonomous indoor and outdoor navigation, content-based image retrieval, etc. With the advances of deep neural networks and image synthesis techniques \cite{Gupta16,jaderberg2014synthetic,zhan2018verisimilar,zhan2019spatial} in recent years, scene text recognition has achieved rapid progress with very impressive recognition performance especially for normal scene texts with clean background and mild distortions \cite{yue2020robustscanner,yu2020towards,shi2018aster}. On the other hand, scene text recognition remains a very challenging task when scene texts are degraded by complex background clutters or severe geometric distortions as illustrated in Fig. \ref{fig:intro}.

Most existing scene text recognition methods can be broadly classified into two categories. The first category follows an encoder-decoder architecture where text images are first encoded into visual features and then directly decoded into a sequence of characters. The \textit{sequential decoding} predicts one character at each time step by focusing on one specific part of image features, which is prone to failure when image features are noisy under the presence of complex background clutters and severe geometric distortions as illustrated in the first column of Fig. \ref{fig:intro}. The failure is largely attributed to the misalignment between visual features and the corresponding decoding time steps. The other category follows a \textit{segmentation} approach that first detects each individual character and then groups the detected characters into words as illustrated in the second column of Fig. \ref{fig:intro}. However, the segmentation approach requires character-level annotations in training which are usually prohibitively expensive to collect. In addition, it often involves various manually-crafted rules in character grouping which are heuristic and prone to errors.

This paper presents I2C2W, an innovative scene text recognition network that takes the merits of \textit{sequential decoding} and \textit{segmentation} but avoids their constraints simultaneously. At one end, it discards the restriction of \textit{time steps} in sequential decoding by first detecting character candidates and then correcting false detection. At the other end, it does not require character-level annotations and the character-to-word mapping is purely learnt without any heuristic rules. These desired features are largely attributed to our designed image-to-character (I2C) and character-to-word (C2W) mappings which are interconnected and end-to-end trainable. Specifically, I2C strives to detect characters candidates in images based on different alignments of visual features which minimizes miss detection at noisy time steps in sequential decoding. It learns to predict relative positions of characters in words (under the supervision of word transcript) without requiring any character-level annotations in training. In addition, C2W learns from the semantics and positions of the detected characters by I2C instead of noisy image features as in \textit{sequential decoding}. It can effectively identify and correct the false character detection that \textit{sequential decoding} cannot handle well.

The contribution of this work is three-fold. First, we design I2C2W, a novel scene text recognition technique that is tolerant to complex background clutters and severe geometric distortions. I2C2W first detects characters and then maps the detected characters to words, which requires no character annotations and heuristic character grouping rules and can handle various false character detection effectively. Second, we design image-to-character and character-to-word mappings which effectively collaborate for predicting character semantics and relative character positions for accurate word recognition. Third, I2C2W is end-to-end trainable and achieves superior recognition performances especially for irregular scene text that suffer from complex background and/or severe geometric distortions, more details to be discussed in Experiments. 

The rest of this paper is organized as follows. Section \ref{sec:related} gives a brief review of related works on scene text recognition and transformer in vision. Section \ref{sec:transformer} introduces the Transformer preliminaries. In Section \ref{sec:method}, we present the proposed I2C2W in details including the character detection, word recognition and network training. Section \ref{sec:experiment} shows experimental settings, quantitative and qualitative experimental results, ablation studies and discussions. Section \ref{sec:conclusion} concludes this paper.

\section{Related Works} \label{sec:related}

\subsection{Scene Text Recognition}
Scene text recognition has been studied for years and most existing works tackle this challenge through either sequential decoding or segmentation.

\noindent \textbf{Scene Text Recognition via Sequential Decoding:} Most recent scene text recognition works follow an encoder-decoder architecture where text images are first encoded into features and then decoded to a sequence of characters directly. Earlier works \cite{jaderberg2014synthetic,jaderberg2016reading} predict word classes directly with encoded image features. As inspired by the advances in NLP, recurrent convolution network (RNN) \cite{elman1990finding} is employed \cite{su2014accurate,he2016reading,shi2016end,bai2018edit} together with a few variants by introducing Aggregation Cross Entropy (ACE) loss \cite{xie2019aggregation} and soft attention \cite{lee2016recursive}. Several attempts \cite{yin2017scene,gao2017reading} also explore to decode character sequences directly from image features by using CNN. In addition, spatial transformer network (STN) \cite{jaderberg2015spatial} with thin-plate spline transformation \cite{bookstein1989principal} has been introduced for the rectification and recognition of irregular texts in scenes \cite{liu2016star,shi2016robust,shi2018aster,zhan2019esir,yang2019symmetry}. Recently, several studies \cite{liu2018squeezedtext,cheng2018aon,li2019show,Litman_2020_CVPR,yang2020holistic,liao2019mask} aim to enhance image encoding and decoding by adopting different attention mechanisms \cite{bahdanau2014neural,xu2015show}. Semantic reasoning \cite{yu2020towards,Fang_2021_CVPR,Qiao_2020_CVPR}, vocabulary \cite{Wan_2020_CVPR}, domain adaptation \cite{zhang2019sequence}, and image super-resolution \cite{yanplugnet,wang2020scene} have also been explored for more accurate scene text recognition. Additionally, recent work \cite{bhunia2021joint,bhunia2021towards,xue2022language} jointly learns the visual and semantic information for better scene text recognition.

\begin{figure*}[!t]
  \centering
  \includegraphics[width=\linewidth]{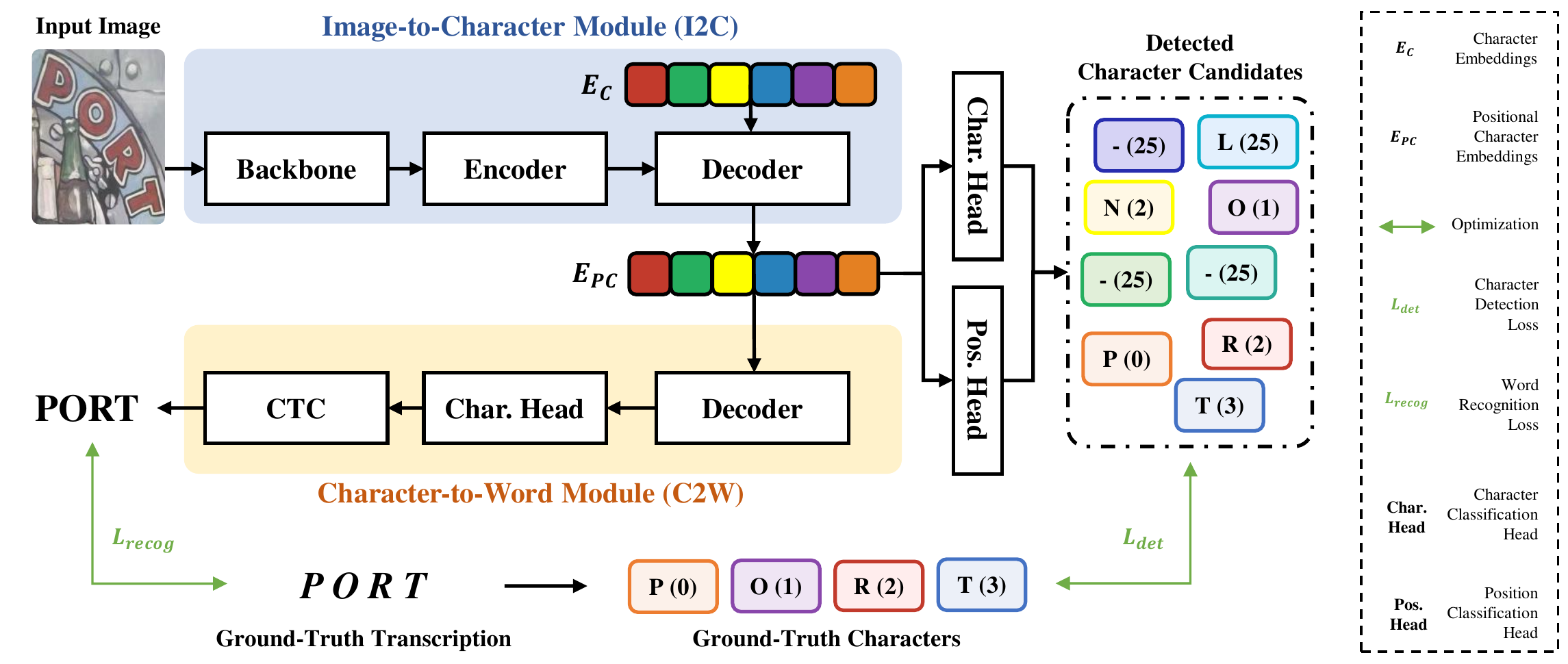}
\caption{
\textbf{The framework of the proposed I2C2W}: Given an \textit{Input Image}, the proposed \textit{Image-to-Character Module} (I2C) first extracts and encodes image features by a \textit{Backbone} and transformer \textit{Encoder}, respectively. It hence takes a set of learnt character embeddings $E_C$ as input queries and predicts positional character embeddings $E_{PC}$ through a transformer \textit{Decoder}. The $E_{PC}$ is mapped to a set of \textit{Detected Character Candidates} by a character classification head (\textit{Char. Head}) and a position classification head (\textit{Pos. Head}) to produce \textit{Detected Character Candidates}. Additionally, the \textit{Character-to-Word Module} (C2W) learns character semantics from $E_{PC}$ and produces the final word recognition result through a transformer \textit{Decoder}, \textit{Char. Head} and \textit{CTC} decoder as illustrated. The I2C2W is optimized by using \textit{Ground-Truth Transcription} and its variant \textit{Ground-Truth Characters} only so it requires no character-level annotations or heuristic character grouping rules in training. }

\label{fig:pipeline}
\end{figure*}

Sequential decoding predicts one character at each time step from the extracted image features. It is sensitive to image noises and often produces missing or redundant predictions if the image features at corresponding time step are noisy. The proposed I2C2W first detects character candidates in images in an non-sequential way which eliminates the time-step restriction but strives to detect all possible character candidates. It learns rich scene text semantics from the detected characters and their positions in words which are essential to scene text recognition.

\noindent \textbf{Scene Text Recognition via Segmentation:} Segmentation-based scene text recognition is inspired by image semantic segmentation, where each pixel is predicted with a semantic class. For example, \cite{wan2020textscanner,liao2019scene,yang2017learning,lyu2018mask} predict character categories at each pixel or detects characters \cite{liu2018char,long2020new}, and then group the predicted characters into words. Segmentation-based recognition requires character-level annotations and involves heuristic rules while grouping the segmented characters into words. The proposed I2C2W does not require character-level annotations as it predicts the relative positions of characters in words instead of their 2D coordinates in images. Its character-to-word grouping is purely learnt from the semantic information of the detected characters which does not involve any heuristic grouping rules.

\subsection{Transformers in Vision}
Transformer \cite{vaswani2017attention} models pairwise interactions between sequence elements which has achieved great success in different NLP tasks \cite{devlin2018bert}. Recently, transformer architecture has been studied intensively in the computer vision research community and its power has been demonstrated in many computer vision tasks. Vision Transformer related systems \cite{dosovitskiy2020image,touvron2021training,yuan2021tokens,jiang2021all,wang2021pyramid} directly apply Transformer on image patches for image classification. DETR and its follow-ups \cite{carion2020end,zhu2020deformable,dai2021up,wang2021anchor,chen2021pix2seq} propose set-based object detectors by leveraging the advances of Transformer. More recently, the Transformer-based network backbone Swin Transformer \cite{liu2021swin} achieves very impressive performances. Additionally, Transformer is applied on other vision tasks such as vision transfer \cite{dosovitskiy2020image}, image GPT \cite{chen2020generative}, image processing \cite{chen2021pre}, image super-resolution \cite{yang2020learning}, colorization \cite{kumar2021colorization}, etc.

The proposed I2C2W consists of two transformer-based networks for mapping from images to characters (I2C) and from characters to words (C2W), respectively. Unlike transformer-based detectors that just localize objects in images, the proposed I2C also predicts relative positions of characters in a word without requiring character-level annotations during training. The proposed C2W recognizes words by learning from character semantics and their positions in words, which is simpler and more robust as compared with sequential decoding that learns to map complex image features to words. 

\section{Transformer Preliminaries}\label{sec:transformer}
The Transformer network usually consists of multi-head attention layers and feed forward networks. 

\noindent \textbf{Attention Layers:} We adopt the attention layers \cite{vaswani2017attention} which are defined by:
\begin{align*}
    \text{Attention}(Q, K, V) = \text{softmax}(\frac{QK^T}{\sqrt{d_k}})V
\end{align*}
where $Q$, $K$ and $V$ refer to input \textit{Queries}, \textit{Keys} and \textit{Values}, respectively. $d_k$ is the dimension of the $Q$ and $K$.

\noindent \textbf{Multi-Head Attention Layers:} The multi-head attention layer \cite{vaswani2017attention} linearly projects queries, keys and values $M$ times by using different learnt linear projections. It is a concatenation of several single attention heads which is defined by:

\begin{figure}[!t]
  \centering
  \includegraphics[width=\linewidth]{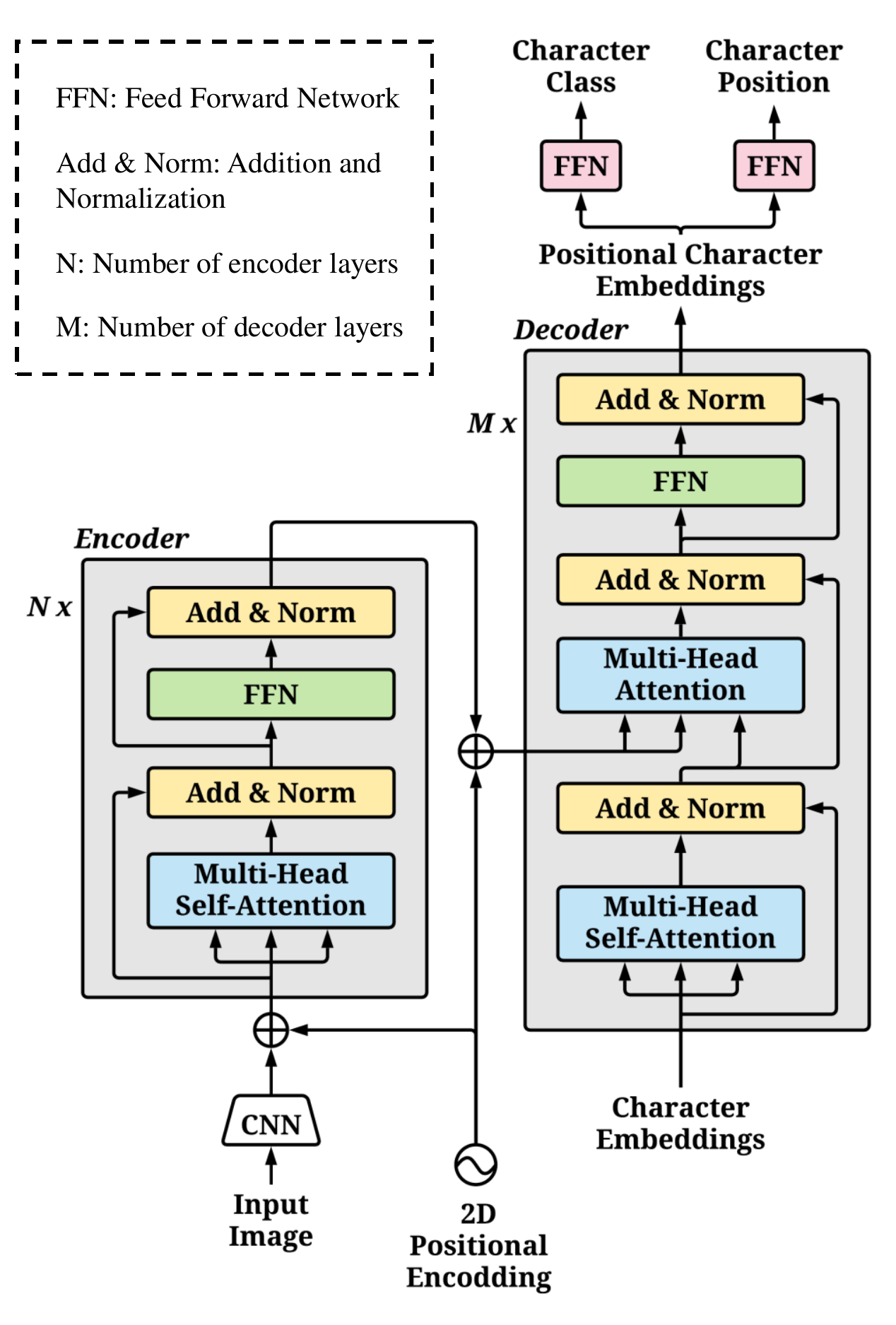}
\caption{The architecture of the proposed Image-to-Character Module.}
\label{fig:sup_net}
\end{figure}

\begin{align*}
  \text{MultiHead}(Q, K, V) = \text{Concat}(head_1, ..., head_M)W^O, \\ 
  \text{where}~head_i = \text{Attention}(QW^Q_i, KW^K_i, VW^V_i)
\end{align*}
where $W^Q_i$, $W^K_i$ and $W^V_i$ are learnt projections of queries, keys and values in head $i$, respectively.

\noindent \textbf{Feed Forward Network (FFN):} The Feed Forward Network consists of two linear transformations with a ReLU activation in between which is defined by:
\begin{align*}
    \text{FFN}(x) = \text{max}(0, xW_1 + b_1)W_2 + b_2
\end{align*}


\section{Methodology}\label{sec:method}

We propose an accurate scene text recognizer I2C2W that formulates scene text recognition as character detection and word recognition tasks as illustrated in Fig. \ref{fig:pipeline}. Given an image, the I2C first predicts a set of \textit{Positional Character Embeddings} that can be mapped to both character categories and character positions as in \textit{Detected Characters Candidates}. The C2W then learns from character semantics with the \textit{Positional Character Embeddings} which predicts words by grouping the characters detected by I2C.

\subsection{Character Detection}

Generic object detectors aim to determine where objects are located in images (localization) and which categories objects belong to (classification). However, generic detection techniques are facing two challenges while tackling character detection in the scene text recognition task. First, they require a large amount of character-level annotations (2D coordinates of characters in images) which are prohibitively time-consuming to collect. Second, they often struggle in grouping the detected characters (with 2D coordinates in images) into words due to the complex shapes and orientations of texts in scenes.

\noindent \textbf{Relative Character Positions:} We re-formulate the character detection problem and propose a novel character detection technique to tackle the aforementioned challenges. The proposed technique decomposes the character detection into two tasks, namely, a classification task that recognizes the category of characters and a localization task that determines the \textit{relative positions} of characters in a word. As illustrated in Fig. \ref{fig:pipeline}, the ground-truth of relative character positions can be directly derived from the word transcription (e.g. `0' and `1' for characters `P' and `O' in the sample word `PORT') and thus character bounding-box annotations are not required during training. At the same time, the relative positions of characters capture the sequence/order of characters in a word which can be employed to learn to achieve character-to-word grouping directly.

We define the relative character position prediction as a (N+1)-category classification problem. Given a word image, the proposed I2C will produce N predictions as illustrated in the \textit{Detected Characters} in Fig. \ref{fig:pipeline}. Here `N’ refers to N categories of character positions, which is a fixed number that is significantly larger than the possible number of characters in a word. The `+1’ refers to a special class `not belongs to words' (as labeled by `25' in Fig. \ref{fig:pipeline}). Similar to the relative position prediction task, we also include an additional class that refers to `not a character' (as labeled by `-' in Fig. \ref{fig:pipeline}) in the character classification task. Under this definition, I2C can learn to separate the image background and foreground characters effectively.

\noindent \textbf{Image-to-Character Mapping:} We adopt the transformer network \cite{vaswani2017attention} in the design of I2C which explicitly models all pairwise interactions between pixels of the whole images. With this nice property, I2C can predict the relative character positions much more effectively as compared with generic object detectors that use CNNs.

Fig. \ref{fig:sup_net} shows the architecture of the proposed I2C. Given an input image $x_{im} \in \mathbb{R}^{3 \times H_0 \times W_0}$, I2C first extracts feature maps $f \in \mathbb{R}^{C \times H \times W}$ and reshapes them into one dimension to produce $z \in \mathbb{R}^{C \times HW}$ which is then fed into the transformer encoder. As the transformer architecture is permutation-invariant, we add a 2D positional encoding (PE) to the input of the attention layer of encoder following \cite{parmar2018image,bello2019attention,carion2020end}. The encoder finally encodes the sequential features into $z_e \in \mathbb{R}^{C \times HW}$.
Note we employ the standard transformer encoder in I2C that includes a multi-head attention layer and a feed forward network (FFN) each of which is followed by a normalization layer.

The standard transformer decoder takes the output from the previous step as input query and makes predictions in a sequential way. Differently, the proposed I2C takes a set of character embeddings $E_{C} \in \mathbb{R}^{D \times N}$ (learnt during training) as input queries and the decoder will predict $N$ characters candidates in a parallel way as illustrated in Fig. \ref{fig:pipeline}. Similar to the encoder, we employ standard transformer decoder that includes two multi-head attention layer and a FFN each of which is followed by a normalization layer. It predicts a set of positional character embeddings $E_{PC} \in \mathbb{R}^{D \times N}$ from the encoded features $z_e$. Finally, the character categories and the relative character positions are obtained from $E_{PC}$ with a character and a positional classification head, respectively.

The proposed I2C aims to resolve the problems of feature misalignment by detecting characters from images. Most existing sequential decoding networks \cite{Fang_2021_CVPR,shi2018aster} decode characters from images in a sequential manner. They are trained to focus on local features at current time steps while often tend to ignore the image features of previous time steps. Therefore, they rely heavily on feature alignment and often fail while image features are not aligned well. Differently, I2C tackles the feature misalignment by detecting characters from images. While detecting a character, it compares local features with the global features without any `time step' restriction. While facing uncertain situations, it considers multiple possible feature alignments and could produce multiple predictions for a single character. The proposed C2W can correct such redundant or false character detection with its learnt linguistic knowledge.

\subsection{Word Recognition}

\textbf{blue}{The detected characters candidates by I2C can be simply grouped into a word according to their predicted relative positions which are usually correct if the text image is clear and clean. However, scene text images often suffer from rich noises due to complex image background and geometric distortions. I2C strives to detect all possible characters and often produces false detection including: 1) incorrect characters due to text-like background patterns (e.g. `L' with position `25' due to the wine bottle in Fig. \ref{fig:pipeline}); 2) redundant detection at one position due to severe distortions or different feature alignments (e.g. `R' and `N' at position `2' in Fig. \ref{fig:pipeline}). We design C2W for robust and accurate word recognition from the noisy character detections by I2C.}

Word recognition aims to predict a sequence of characters that belong to a word. Most existing methods decode the sequence of characters from the original image features directly which often fail when the visual features are too noisy to be aligned to the corresponding \textit{time steps} correctly. Instead of recognizing words from noisy image features, the proposed C2W predicts a word from a set of `noisy characters' (with character semantics and their relative positions) as detected by I2C which is much simpler and robust than sequential decoding from visual image features.

\noindent \textbf{Character-to-Word Mapping:} The proposed C2W aims to correct falsely detected characters by learning from character semantics and their relative positions from I2C. We adopt a transformer network that has been proved effective in learning pairwise relations of words in sentences in various NLP tasks. In the C2W, we treat characters as the smallest elements and learn the semantics of characters in a word for accurate word recognition.

\begin{figure}[!t]
  \centering
  \includegraphics[width=\linewidth]{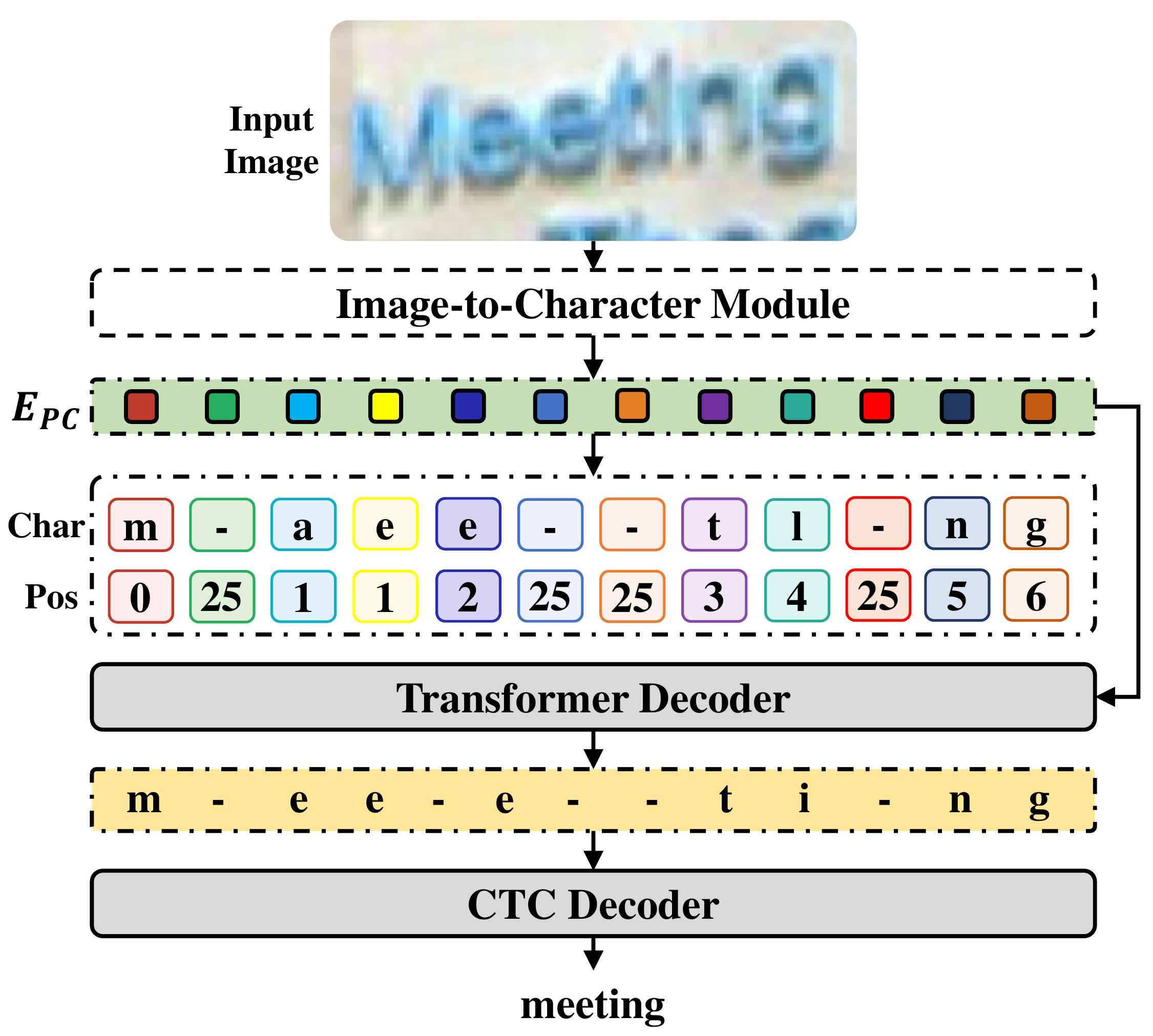}
\caption{\textbf{Illustration of the proposed character-to-word module (C2W)}: Given an input image, I2C may detect redundant or wrong characters (e.g. characters `a' and `l'), which can be corrected by the proposed C2W effectively. The CTC decoder is adopted to produce the final recognition. Note the detected characters are sorted for visualization only. Char: Character; Pos: Relative Position. $E_{PC}$: Positional Character Embeddings. `-': Not a character. `25': Not belongs to word.}
\label{fig:c2w}
\end{figure}

Similar to I2C, we adopt standard transformer decoder in C2W and use learnt embedding as queries. Different from existing methods, we take the \textit{Positional Character Embeddings} as the input (instead of mapping I2C detected characters to embeddings) due to two factors. First, \textit{Positional Character Embeddings} are a combination of character embeddings and positional encoding as they are learnt by being mapped to character classes and positions. They improve the efficiency of I2C2W by excluding redundant character embeddings and positional encoding steps. Second, \textit{Positional Character Embeddings} contain detailed character semantic information (e.g. the visual probabilities of each class for each character) that is helpful in word recognition. For example, if a character sequence `hcat' is detected and both `h' and `c' are at position `0', it would be difficult for C2W to tell if the word is `hat' or `cat' as both are valid words. By taking the \textit{Positional Character Embeddings} as input, C2W could produce correct recognition by considering the visual probabilities of `h' and `c'. In this way, I2C and C2W could complement each other effectively.

Fig. \ref{fig:c2w} shows the word recognition pipeline of the proposed C2W. With the input $E_{PC} \in \mathbb{R}^{D \times N}$, C2W outputs a sequence of characters by correcting false detection (i.e. `a' $\rightarrow$ `e' and `l' $\rightarrow$ `i' in Fig. \ref{fig:c2w}). Note that the C2W takes a set of learnt embeddings (initialized randomly) as queries instead of position encodings in \cite{yu2020towards,Fang_2021_CVPR}. It hence corrects each false detection from I2C and may produces duplicated characters at one relative positions (i.e. duplicated `e's at position `1' in Fig. \ref{fig:c2w}). We adopt a Connectionist Temporal Classification (CTC) decoder \cite{graves2006connectionist} to remove duplicated characters at the same relative position and produce final recognition $\textbf{y}$. This design enables the C2W to correct false I2C recognition I2C of different cases, including words with missing characters (by correcting `-' to correct character), redundant characters (by correcting wrong character to `-') and wrong characters (by correcting wrong characters to correct ones).

\subsection{Network Training}
The training of the proposed I2C2W aims to minimize the following multi-task loss function:
\begin{equation}
\mathcal{L}=\mathcal{L}_{det} + \mathcal{L}_{recog}
\end{equation}
\noindent where $\mathcal{L}_{det}$ and $\mathcal{L}_{recog}$ refer to the loss of character detection and word recognition tasks, respectively.

\noindent \textbf{Character Detection:} The I2C outputs a predicted character set $\hat{y}$ which contains character semantics and position information. We first find a bipartite matching between predicted character set $\hat{y}$ and ground-truth character set $y$ following \cite{carion2020end} with the lowest cost. Different from \cite{carion2020end}, the matching cost takes both character semantics and character positions into account which is computed by:
\begin{equation}
\mathcal{L}_{match}(y_i, \hat{y}_i)= \hat{C}_{\theta(i)}(c_i) + \beta \hat{L}_{\theta(i)}(l_i)
\end{equation}
The detection loss $\mathcal{L}_{det}$ is hence computed by:

\begin{equation}
\mathcal{L}_{det}= \mathcal{L}_{CE}(\hat{y}^{c}_{\hat{\theta}(i)}, y^{c}_{i}) + \mathcal{L}_{CE}(\hat{y}^{l}_{\hat{\theta}(i)}, y^{l}_{i})
\end{equation}
\noindent where $(\hat{\theta}(i), i)$ are indices of matched pairs in the predicted and ground-truth sets, respectively. $c$ and $l$ refer to the character semantics and character position, respectively. $\mathcal{L}_{CE}$ is the standard cross-entropy loss. In practice, the losses of `not a character' and `not belongs to word' categories are down-weighted with factor of 10 to balance the losses. 

\noindent \textbf{Word Recognition:} As the detected character sequence contains a large number of predictions of `not a character', `not belongs to word' and multiple redundant detection (at the same position), we adopt the CTC loss \cite{graves2006connectionist} for word recognition. The objective is to minimize negative log-likelihood of conditional probability of ground truth:
\begin{equation}
\mathcal{L}_{recog} = - \text{log}\:p(\textbf{l}|\textbf{y}),
\end{equation}
where $l$ and $y$ refer to the ground-truth transcription and recognized character sequence, respectively. The conditional probability is defined by:
\begin{equation}
p(\textbf{l}|\textbf{y}) = \sum_{\pi \in B^{-1}(l)}\:p(\pi|\textbf{y}), 
\end{equation}
\begin{equation}
p(\pi|\textbf{y}) = \prod y_{\pi_t}^t,
\end{equation}
where the $B$ is the mapping function and $B^{-1}(l)$ refers to the set of all posible sequences $\pi$ which outputs $l$ from $B$.

\begin{table}[!t]
\centering
\caption{The details of 11 public datasets that are used for training and testing in the experiments.}
\begin{tabular}{cccc}
\toprule
Type                         & Dataset  & \#Train & \#Test \\
\midrule
\multirow{2}{*}{Synthetic}   & ST       & 7M     & -     \\
                             & 90K      & 9M     & -     \\
\midrule
\multirow{3}{*}{Curved}      & CUTE     & -      & 228   \\
                             & Total    & -      & 2,209 \\
                             & SCUT-CTW & -      & 5,040 \\
\midrule 
\multirow{3}{*}{Multi-oriented} & IC15     & 4,468  & 1,811 \\
                             & SVT      & -      & 647   \\
                             & SVTP     & -      & 645   \\
\midrule 
\multirow{3}{*}{Normal}      & IIIT5K   & 2,000  & 3,000 \\
                             & IC03     & -      & 860   \\
                             & IC13     & 848    & 1,095 \\
\bottomrule                             
\end{tabular}
\label{tab:dataset}
\end{table}

\begin{table*}[!t]
\caption{Scene text recognition performance over \textbf{curved datasets}: All results are obtained without lexicon. The used training data and annotations are shown in columns `Data' and `Annotation', where `Real' and `word' refer to real-world dataset and word-level annotation, respectively. The first section shows the methods that use different training settings. The second and third sections report the models that use ResNet (as backbone) without/with using real data in training, respectively.}
\centering
\begin{tabular}{llccccc}
\toprule
\multirow{2}{*}{\textbf{Method}}                                & \multirow{2}{*}{\textbf{Data}} & \multirow{2}{*}{\textbf{Annotation}} & \multirow{2}{*}{\textbf{Backbone}} & \multicolumn{3}{c}{\textbf{Curved Dataset}}              \\ \cline{5-7} 
                                                                &                                &                                 &                                    & \textbf{CUTE} & \textbf{TOTAL} & \textbf{SCUT-CTW} \\ 
\midrule                                                        
Shi \textit{et al.} (CTC) \cite{shi2016end}                     & Synth90K                            & word                            & VGG                                & 60.1          & 52.1           & 59.9              \\
Zhang \textit{et al.} (AutoSTR) \cite{zhang2020autostr}         & Synth90K+SynthText                         & word                            & NAS                                & 82.6          & 74.7           & 74.5              \\ \midrule
Shi \textit{et al.} (ASTER) \cite{shi2018aster}                 & Synth90K+SynthText                         & word                            & ResNet                             & 79.5          & 66.8           & 66.1              \\
Yu \textit{et al.} (SRN) \cite{yu2020towards}                    & Synth90K+SynthText                         & word                            & ResNet                             & 87.8          & 68.9           & 72.6              \\
\textbf{Ours (I2C2W)}                                           & Synth90K+SynthText                         & word                            & ResNet                             & \textbf{93.1} & \textbf{79.8}  & \textbf{79.1}     \\ \midrule
Li \textit{et al.} (SAR) \cite{li2019show}                      & Synth90K+SynthText+Real                       & word                            & ResNet                             & 89.6          & 78.4           & 78.3              \\
Yue \textit{et al.} (RobustScanner) \cite{yue2020robustscanner} & Synth90K+SynthText+Real                       & word                            & ResNet                             & 92.4          & 77.8           & 78.2              \\
\textbf{Ours (I2C2W)}                                           & Synth90K+SynthText+Real                       & word                            & ResNet                             & \textbf{93.1} & \textbf{81.5}  & \textbf{79.7}     \\ 
\bottomrule
\end{tabular}
\label{tab:comparison_curve}
\end{table*}

\section{Experiments}\label{sec:experiment}
We present experimental results in this section. Specifically, Sections \ref{sec:dataset} and \ref{sec:metrics} describe the details of datasets and evaluation metrics, respectively. Section \ref{sec:implementation} presents the implementation details of the proposed I2C2W. Sections \ref{sec:comparison} and \ref{sec:ablation} present comparisons with the state-of-the-art and ablation studies, respectively. Finally, Section \ref{sec:discussion} discusses some unique features of the proposed I2C2W.

\subsection{Datasets}\label{sec:dataset}
We evaluate I2C2W and benchmark it with the state-of-the-art over three sets of public datasets that have been widely used in the scene text recognition study as shown in Table \ref{tab:dataset}. In addition, we also use two widely used synthetic datasets for network training.

\noindent \textbf{Synthetic Datasets:} For fair comparison, we follow \cite{baek2019wrong} and adopt two widely adopted synthetic datasets to train scene text recognition model.

\noindent 1) \textit{SynthText} (SynthText) \cite{jaderberg2014synthetic} that has been widely used for scene text recognition research as well by cropping text image patches according to the provided text annotation boxes. We crop 7 million text image patches from this dataset for training.

\noindent 2) \textit{Synth90K} (Synth90K) \cite{gupta2016synthetic} contains 9 million synthetic text images which have been widely used for training scene text recognition models. It has no separation of training and test data and all images are used for training.

\noindent \textbf{Curved Datasets:} We evaluate and benchmark I2C2W over three challenging curved datasets, where many texts in images suffer from severe geometric distortions or are with arbitrary shapes and complex background.

\noindent 1) \textit{CUTE80} (CUTE) \cite{risnumawan2014robust} consists of 288 word images where most scene texts are curved. All text images are cropped from CUTE datasets with 80 scene text images.

\noindent 2) \textit{Total-Text} (TOTAL) \cite{ch2017total} contains 1,253 training images and 300 test images which have been widely used for the research of arbitrary-shaped scene text detection. We crop 2,209 word images from the test set of Total-Text dataset by using the provided annotation boxes.

\noindent 3) \textit{SCUT-CTW1500} (SCUT-CTW) \cite{liu2019curved} consists of 1,000 training images and 500 test images. We crop 5,040 word images from the test set of SCUT-CTW dataset by using the provided word-level annotation boxes.

\noindent \textbf{Multi-oriented Datasets:} We also evaluate I2C2W over three multi-oriented datasets where many scene texts suffer from clear perspective distortions.

\noindent 1) \textit{ICDAR-2015} (IC15) \cite{karatzas2015icdar} contains incidental scene text images that are captured without preparation before capturing. It contains 4,468 text patches for training and 1,811 patches for test which are cropped from the original dataset.

\noindent 2) \textit{Street View Text} (SVT) \cite{wang2011end} consists of 647 word images that are cropped from 249 street view images from Google Street View and most cropped word images are almost horizontal.

\noindent 3) \textit{Street View Text-Perspective} (SVTP) \cite{phan2013recognizing} contains 645 word images that are also cropped from Google Street View and many of them suffer from perspective distortions.

\noindent \textbf{Normal Datasets:} We also evaluated and benchmark I2C2W over three widely used normal datasets where most texts in scene images are horizontal with fair image quality.

\noindent 1) \textit{IIIT 5K-words} (IIIT5K) \cite{mishra2012top} contains 2,000 training and 3,000 test word patches cropped from born-digital images collected from Google image search.

\noindent 2) \textit{ICDAR-2003} (IC03) \cite{lucas2005icdar} contains 860 images of cropped word from the the Robust Reading Competition in the International Conference on Document Analysis and Recognition (ICDAR) 2003 dataset.

\noindent 3) \textit{ICDAR-2013} (IC13) \cite{karatzas2013icdar} is used in the Robust Reading Competition in the ICDAR 2013 which contains 848 word images for training and 1,095 for testing.

\subsection{Evaluation Metrics}\label{sec:metrics}
Following recent scene text recognition work \cite{yue2020robustscanner,zhan2019esir,baek2019wrong}, we evaluate the proposed I2C2W in terms of accuracy. Specifically, a success sample is counted if the word in an image is recognized correctly. We compute the success rate of word predictions over the images in each public dataset.

\subsection{Implementation Details}\label{sec:implementation}
We adopt ResNet-50 (pretrained on ImageNet) as CNN backbone and optimize I2C2W by AdamW optimizer with batch size of 48. The learning rate is set to $10^{-5}$ for backbone and  $10^{-4}$ for transformers. The numbers of the transformer encoder/decoder layers of I2C and C2W are set at 3 and 1, respectively. We apply dropout of 0.1 for every multi-head attention and FFN layers before the normalization layers. The weights of I2C and C2W are randomly initialized with Xavier initialization. I2C2W is trained end-to-end and all experiments are implemented with 4 Telsa V100 GPUs. The shorter side of input images are randomly resized to a number in $(32, 96)$ and the size of the longer side is computed according to the original aspect ratio but cap at 600. Different data augmentation operations are adopted following \cite{yu2020towards,Fang_2021_CVPR}. The number of class is 37 including 0-9, a-z, and `not a character’, and the 25 characters in total are detected in I2C following \cite{yu2020towards,Fang_2021_CVPR}. In the test stage, the shorter side of input images is resized to 64 and the longer side is computed based on the original aspect ratio. 

\begin{table*}[!t]
\caption{Scene text recognition performance over \textbf{multi-oriented datasets}: All results are obtained without lexicon. The training data and annotations are shown in columns `Data' and `Annotation', where `SynthAdd', `Lang', `Real', `word' and `char' refer to synthetic dataset from \cite{li2019show}, language dataset, real-world dataset, word-level annotation and character-level annotation, respectively. The first section shows the methods that use different training settings. The second and third sections report the models that use ResNet without/with using real data in training, respectively.}
\centering
\begin{tabular}{llccccc}
\toprule
\multirow{2}{*}{\textbf{Method}}                                & \multirow{2}{*}{\textbf{Data}} & \multirow{2}{*}{\textbf{Annotation}} & \multirow{2}{*}{\textbf{Backbone}} & \multicolumn{3}{c}{\textbf{Multi-oriented Dataset}}                                                           \\ \cline{5-7} 
                                                                &                                &                                 &                                    & \textbf{IC15} & \textbf{SVT}  & \textbf{SVTP} \\ \midrule
Xie \textit{et al.} (ACE) \cite{xie2019aggregation}             & Synth90K                            & word                            & ResNet                               & 68.9          & 82.6          & 70.1          \\
Liao \textit{et al.} (FCN) \cite{liao2019scene}                 & SynthText                             & word+char                       & VGG                                 & -             & 86.4          & -             \\
Wan \textit{et al.} (2D-CTC) \cite{wan20192d}                   & Synth90K+SynthText                         & word                            & PSPNet                              & 75.2          & 90.6          & 79.2          \\
Yang \textit{et al.} (ScRN) \cite{yang2019symmetry}             & Synth90K+SynthText                         & word+char                       & ResNet                              & 78.7          & 88.9          & 80.8          \\
Zhang \textit{et al.} (AutoSTR) \cite{zhang2020autostr}         & Synth90K+SynthText                         & word                            & NAS                                  & 81.8          & 90.9          & 81.7          \\ 
Litman \textit{et al.} (SCATTER) \cite{Litman_2020_CVPR}        & Synth90K+SynthText+SynthAdd                      & word                            & ResNet                               & 82.2          & 92.7          & 86.9          \\
Qiao \textit{et al.} (SE-ASTER) \cite{Qiao_2020_CVPR}           & Synth90K+SynthText+Lang                       & word                            & ResNet                              & 80.0          & 89.6          & 81.4          \\
Yan \textit{et al.} (PREN2D) \cite{Yan_2021_CVPR}               & Synth90K+SynthText                         & word                            & EfficientNet                         & 83.0          & 94.0          & 87.0          \\
Fang \textit{et al.} (ABINet) \cite{Fang_2021_CVPR}             & Synth90K+SynthText+Lang                       & word                            & ResNet                               & 86.0          & 93.5          & 89.3          \\\midrule
Shi \textit{et al.} (ASTER) \cite{shi2018aster}                 & Synth90K+SynthText                         & word                            & ResNet                              & 76.1          & 89.5          & 78.5          \\
Lyu \textit{et al.} (Parallel) \cite{lyu20192d}                 & Synth90K+SynthText                         & word                            & ResNet                              & 76.3          & 90.1          & 82.3          \\
Zhan \textit{et al.} (ESIR) \cite{zhan2019esir}                 & Synth90K+SynthText                         & word                            & ResNet                              & 76.9          & 90.2          & 79.6          \\
Mou \textit{et al.} (PlugNet) \cite{yanplugnet}                 & Synth90K+SynthText                         & word                            & ResNet                              & 82.2          & \textbf{92.3} & 84.3          \\
Yu \textit{et al.} (SRN) \cite{yu2020towards}                    & Synth90K+SynthText                         & word                            & ResNet                              & 82.7          & 91.5          & \textbf{85.1} \\

Bhunia \textit{et al.} \cite{bhunia2021joint} & Synth90K+SynthText                         & word                            & ResNet                              & \textbf{84.0}          & 92.2          & 84.7\\
\textbf{Ours (I2C2W)}                                           & Synth90K+SynthText                         & word                            & ResNet                              & 82.8 & 91.7          & 83.1          \\ \midrule
Li \textit{et al.} (SAR) \cite{li2019show}                      & Synth90K+SynthText+Real                       & word                            & ResNet                               & 78.8          & 91.2          & \textbf{86.4}          \\
Yue \textit{et al.} (RobustScanner) \cite{yue2020robustscanner} & Synth90K+SynthText+Real                       & word                            & ResNet                                   & 79.2          & 89.3          & 82.9          \\
\textbf{Ours (I2C2W)}                                           & Synth90K+SynthText+Real                       & word                            & ResNet                              & \textbf{85.3} & \textbf{92.8} & 84.7          \\ \bottomrule
\end{tabular}
\label{tab:comparison_perspective}
\end{table*}

\begin{table*}[!t]
\caption{Scene text recognition performance over \textbf{normal datasets}: All results are obtained without lexicon. The training data and annotations are shown in columns `Data' and `Annotation', where `SynthAdd', `Lang', `Real', `word' and `char' refer to synthetic dataset from \cite{li2019show}, Language dataset, real-world dataset, word-level annotation and character-level annotation, respectively. The first section shows the methods that use different training settings. The second and third sections report the models that use ResNet without/with using real data in training, respectively.}
\centering
\begin{tabular}{llccccc}
\toprule
\multirow{2}{*}{\textbf{Method}}                                & \multirow{2}{*}{\textbf{Data}} & \multirow{2}{*}{\textbf{Annotation}} & \multirow{2}{*}{\textbf{Backbone}} & \multicolumn{3}{c}{\textbf{Normal Dataset}}                                                           \\ \cline{5-7} 
                                                                &                                &                                 &                                    & \textbf{IIIT5K} & \textbf{IC03} & \textbf{IC13} \\ \midrule
Xie \textit{et al.} (ACE) \cite{xie2019aggregation}             & Synth90K                            & word                            & ResNet                             & 82.3            & 92.1          & 89.7                    \\
Liao \textit{et al.} (FCN) \cite{liao2019scene}                 & SynthText                             & word+char                       & VGG                                & 91.9            & -             & 91.5                  \\
Wan \textit{et al.} (2D-CTC) \cite{wan20192d}                   & Synth90K+SynthText                         & word                            & PSPNet                             & 94.7            & -             & 93.9                 \\
Yang \textit{et al.} (ScRN) \cite{yang2019symmetry}             & Synth90K+SynthText                         & word+char                       & ResNet                             & 94.4            & 95.0          & 93.9                    \\
Zhang \textit{et al.} (AutoSTR) \cite{zhang2020autostr}         & Synth90K+SynthText                         & word                            & NAS                                & 94.7            & 93.3          & 94.2                   \\ 
Litman \textit{et al.} (SCATTER) \cite{Litman_2020_CVPR}        & Synth90K+SynthText+SynthAdd                      & word                            & ResNet                             & 93.7            & 96.3          & 93.9                  \\
Qiao \textit{et al.} (SE-ASTER) \cite{Qiao_2020_CVPR}           & Synth90K+SynthText+Lang                       & word                            & ResNet                             & 93.8            & -             & 92.8                 \\
Yan \textit{et al.} (PREN2D) \cite{Yan_2021_CVPR}               & Synth90K+SynthText                         & word                            & EfficientNet                       & 95.6            & 95.8          & 96.4                 \\
Fang \textit{et al.} (ABINet) \cite{Fang_2021_CVPR}             & Synth90K+SynthText+Lang                       & word                            & ResNet                             & -               & -             & 97.4                   \\\midrule
Shi \textit{et al.} (ASTER) \cite{shi2018aster}                 & Synth90K+SynthText                         & word                            & ResNet                             & 93.4            & 94.5          & 91.5                  \\
Lyu \textit{et al.} (Parallel) \cite{lyu20192d}                 & Synth90K+SynthText                         & word                            & ResNet                             & 94.0            & 94.3          & 92.7                \\
Zhan \textit{et al.} (ESIR) \cite{zhan2019esir}                 & Synth90K+SynthText                         & word                            & ResNet                             & 93.3            & -             & 91.3                 \\
Mou \textit{et al.} (PlugNet) \cite{yanplugnet}                 & Synth90K+SynthText                         & word                            & ResNet                             & 94.4            & \textbf{95.7}          & 95.0            \\
Yu \textit{et al.} (SRN) \cite{yu2020towards}                    & Synth90K+SynthText                         & word                            & ResNet                             & 94.8            & 94.6          & \textbf{95.5}  \\
Bhunia \textit{et al.} \cite{bhunia2021joint} & Synth90K+SynthText                         & word                            & ResNet                              & \textbf{95.2}          & -          & \textbf{95.5} \\
\textbf{Ours (I2C2W)}                                           & Synth90K+SynthText                         & word                            & ResNet                             & 94.3            & 95.2          & 95.0                \\ \midrule
Li \textit{et al.} (SAR) \cite{li2019show}                      & Synth90K+SynthText+Real                       & word                            & ResNet                             & 95.0            & 93.6          & 94.0                  \\
Yue \textit{et al.} (RobustScanner) \cite{yue2020robustscanner} & Synth90K+SynthText+Real                       & word                            & ResNet                             & \textbf{95.4}            & 93.7          & 94.1                   \\
\textbf{Ours (I2C2W)}                                           & Synth90K+SynthText+Real                       & word                            & ResNet                             & 94.5        &  \textbf{95.1}        & \textbf{94.8}         \\ \bottomrule
\end{tabular}
\label{tab:comparison_normal}
\end{table*}

\begin{figure*}[!t]
  \centering
  \includegraphics[width=\linewidth]{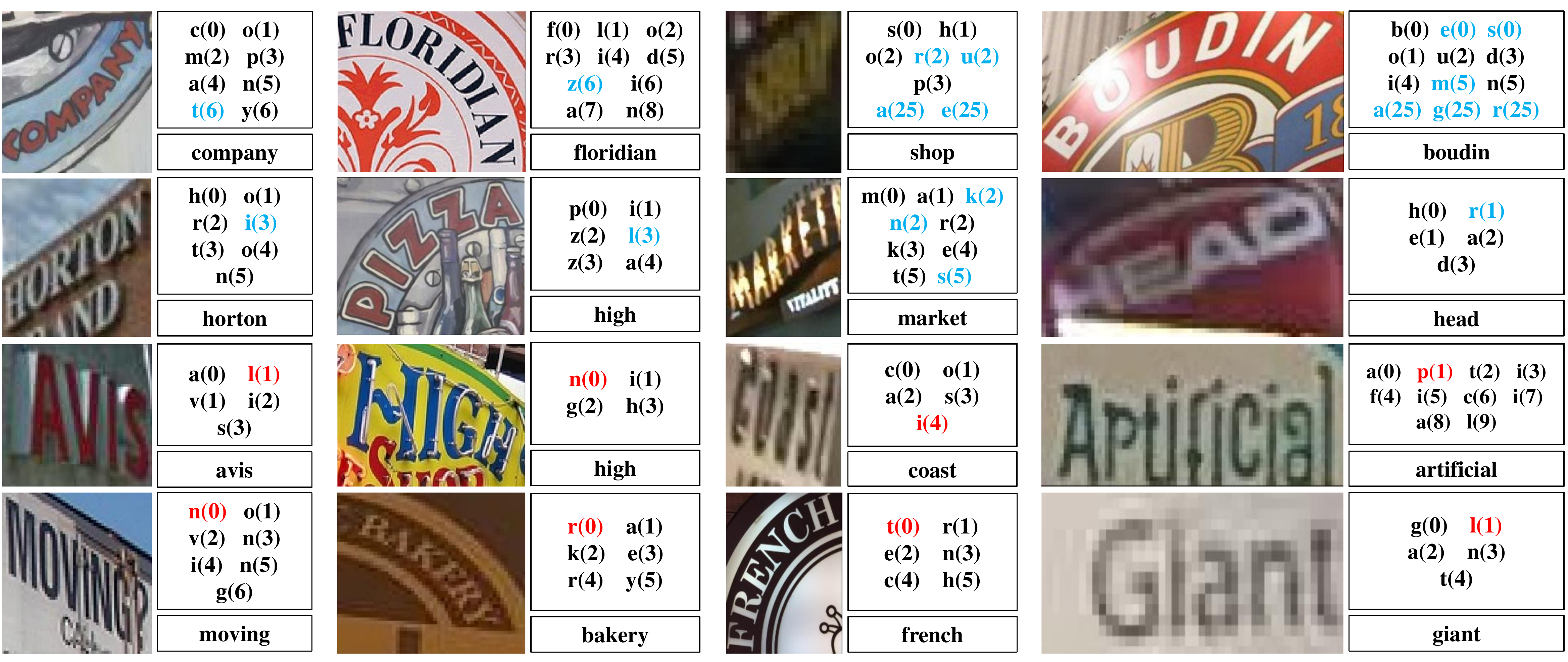}
\caption{
\textbf{Illustration of scene text recognition by the proposed I2C2W}: For each sample image from the public datasets (TOTAL, CTW, IC15, SVT, and SVTP), the graph on the right shows the recognition by I2C2W. Specifically, the upper part of the graph shows the detected character candidates by I2C, where the relative character positions are included in (). The redundant and incorrect character detection is highlighted by blue and red colors, respectively. The lower part of the graph shows the recognized word. The proposed I2C and C2W successfully detect characters in images and refines them to produce more accurate word recognition, respectively.
}
\label{fig:ablation}
\end{figure*}

\subsection{Comparisons with State-of-the-Art Methods}\label{sec:comparison}

We evaluate I2C2W extensively over the nine datasets as described in Section \ref{sec:dataset} Dataset. For each of the nine datasets, we conduct two types of experiments following two most widely-adopted training settings for fair comparisons with the state-of-the-art. The first setting trains scene text recognition models by using the two synthetic datasets only, and then evaluate the trained models over the test images of the nine datasets. The second setting further fine-tunes the models (trained using the synthetic images under the first setting) by using training images of relevant real datasets as in \cite{yue2020robustscanner,li2019show}. Tables \ref{tab:comparison_curve}-\ref{tab:comparison_normal} show the corresponding quantitative scene text recognition accuracy of the three sets of datasets, respectively. Fig. \ref{fig:ablation} illustrates the scene text detection by the proposed I2C2W.

\noindent\textbf{Recognizing Curved Text:} We compare I2C2W with the state-of-the-art over the three curved datasets CUTE, TOTAL and SCUT-CTW where scene text images suffer from clear geometric distortions. Table \ref{tab:comparison_curve} shows experimental results. It can be seen clearly that  I2C2W outperforms the state-of-the-art by 5.3, 10.9 and 6.5 on CUTE, TOTAL and SCUT-CTW, respectively,  while using synthetic data only in  training. While fine-tuning the trained models on the training images of three curved datasets, I2C2W outperforms the state-of-the-art by 0.7, 3.7 and 1.5, respectively, for the three curved datasets as shown in the third section of Table \ref{tab:comparison_curve}. In particular, many images in the three curved datasets suffer from severe geometric distortions, arbitrary shapes or complex and noisy image background which often lead to incorrect alignments of visual features at noisy \textit{time steps} as in existing methods. The proposed I2C2W discards the time-step concept by first considering multiple visual alignments for character detection and then learning to map the detected characters to words. It is thus more tolerant to text shape variations and complex image background.  

\noindent\textbf{Recognizing Multi-Oriented Text:} We benchmark I2C2W with the state-of-the-art over the three multi-oriented datasets as well, including IC15, SVT and SVTP datasets where most texts are multi-oriented with perspective distortions. The three datasets have been widely investigated with different training data, backbone networks, as well as sophisticated optimization techniques. We therefore grouped the benchmarking into two categories, where the first category includes existing methods that use different setups of training data and/or backbone networks (as shown in the first section of Table \ref{tab:comparison_perspective}) and the second category includes existing methods that adopt similar training data and backbone networks (as in the second and third section). As Table \ref{tab:comparison_perspective} shows, the simple implementation of I2C2W with little optimization can achieve state-of-the-art recognition accuracy over scene texts with various perspective distortions.

\noindent\textbf{Recognizing Normal Text:} We also benchmark the proposed I2C2W with the state-of-the-art over the three normal datasets including IIIT5K, IC03, IC13. The three datasets have been studied for years under different settings in training data, backbone networks, and optimization techniques. Similar to the benchmarking with the three multi-oriented datasets, we group existing methods into two categories where methods in the first category use additional training data and/or more advanced backbone networks (as shown in the first section of Table \ref{tab:comparison_normal}) and methods in the second category use similar training data and backbone networks as I2C2W (as shown in the second and third sections). As Table \ref{tab:comparison_normal} shows, the simply implementation of the proposed I2C2W achieves state-of-the-art recognition accuracy for the widely-existed and long-studied scene texts with normal geometry and appearance.

\begin{figure*}[h]
  \centering
  \includegraphics[width=\linewidth]{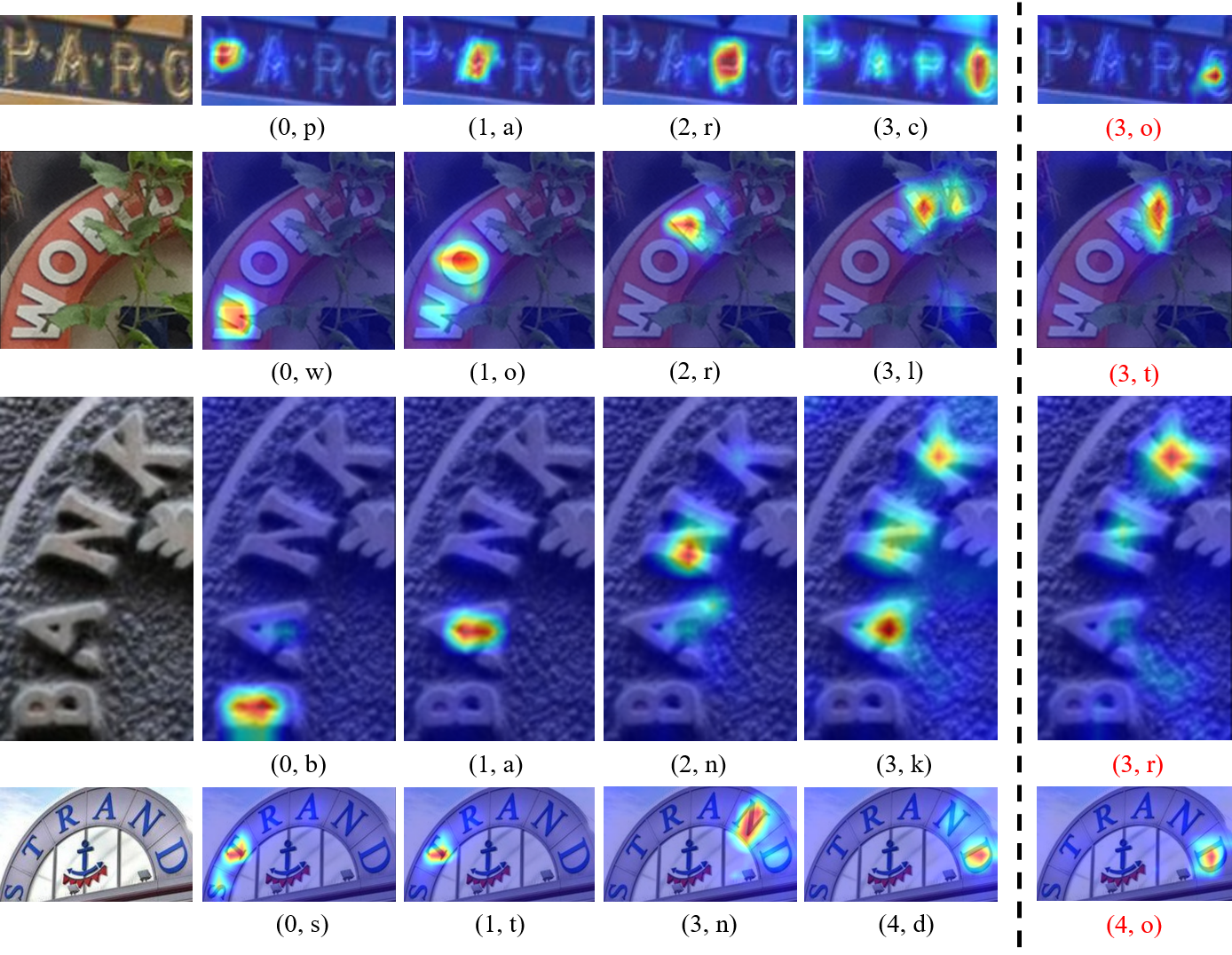}
\caption{
Illustration of attention maps in character detection in I2C2W: With sample images in the first columns, columns 2-5 show the character attention maps as produced by I2C2W. The last column shows the attention maps of false detection by I2C only. The predicted character classes and relative character positions are provided in () at the bottom of each image. It can be seen that I2C2W focuses on character regions and detect character accurately as shown in columns 2-5. With C2W, I2C can learn to focus on neighbouring character regions to produce more accurate detection of noisy characters as shown in the second last column instead of just focusing on local features as by using the I2C only as illustrated in the last column.
}
\label{fig:sample}
\end{figure*}

\begin{table}[!t]
\centering
\caption{\textbf{Ablation studies on curved dataset}: The accuracy of the proposed I2C and I2C2W on curved datasets are shown on the two rows, respectively. `S' and `R' refer to synthetic dataset and real-world dataset, respectively}
\begin{tabular}{ccccccc}
\toprule
\multirow{2}{*}{\textbf{Model}} & \multicolumn{2}{c}{\textbf{CUTE}} & \multicolumn{2}{c}{\textbf{TOTAL}} & \multicolumn{2}{c}{\textbf{SCUT-CTW}} \\ \cmidrule(lr){2-3}\cmidrule(lr){4-5}\cmidrule(lr){6-7}
                                & \textbf{S}     & \textbf{S+R}    & \textbf{S}      & \textbf{S+R}     & \textbf{S}       & \textbf{S+R}       \\
\midrule
I2C                    & 81.2       & 82.5       & 71.0        & 71.7        & 72.4          & 74.5         \\
I2C2W                  & 93.1       & 93.1       & 79.8        & 80.9        & 79.1          & 79.7        \\
\bottomrule
\end{tabular}
\label{tab:ablation_curved}
\end{table}

\subsection{Ablation Studies}\label{sec:ablation}

We also investigated how the proposed I2C and C2W contribute to the overall scene text recognition accuracy with extensive ablation experiments over the nine public datasets. Specifically, we evaluate the I2C recognition performance by ordering the I2C-detected character candidates according to the predicted character positions and further filtering out the character candidates with a label of `not a character' or `not belongs to word'. The contribution of C2W to the overall recognition accuracy can thus be inferred from the difference of accuracy of the I2C2W models and the I2C models. Tables \ref{tab:ablation_curved}-\ref{tab:ablation_normal} show quantitative ablation study experiments, and Fig. \ref{fig:ablation} further shows qualitative illustrations on how I2C and C2W contribute to the overall recognition accuracy.

\begin{table}[!t]
\centering
\caption{\textbf{Ablation studies on multi-oriented dataset}: The accuracy of the proposed I2C and I2C2W on multi-oriented datasets are shown on the two rows, respectively. `S' and `R' refer to synthetic dataset and real-world dataset, respectively}
\begin{tabular}{ccccccc}
\toprule
\multirow{2}{*}{\textbf{Model}} & \multicolumn{2}{c}{\textbf{IC15}} & \multicolumn{2}{c}{\textbf{SVT}} & \multicolumn{2}{c}{\textbf{SVTP}} \\ \cmidrule(lr){2-3}\cmidrule(lr){4-5}\cmidrule(lr){6-7}
                                & \textbf{S}     & \textbf{S+R}    & \textbf{S}      & \textbf{S+R}     & \textbf{S}       & \textbf{S+R}       \\
\midrule
I2C                    & 75.2       & 76.7       & 81.9        & 82.9        & 72.1          & 73.2         \\
I2C2W                  & 82.8       & 85.3       & 91.7        & 92.8        & 83.1          & 84.7        \\
\bottomrule
\end{tabular}
\label{tab:ablation_perspective}
\end{table}

\begin{table}[!t]
\centering
\caption{\textbf{Ablation studies on normal dataset}: The accuracy of the proposed I2C and I2C2W on normal datasets are shown on the two rows, respectively. `S' and `R' refer to synthetic dataset and real-world dataset, respectively}
\begin{tabular}{ccccccc}
\toprule
\multirow{2}{*}{\textbf{Model}} & \multicolumn{2}{c}{\textbf{IIIT5K}} & \multicolumn{2}{c}{\textbf{IC03}} & \multicolumn{2}{c}{\textbf{IC13}} \\ \cmidrule(lr){2-3}\cmidrule(lr){4-5}\cmidrule(lr){6-7}
                                & \textbf{S}     & \textbf{S+R}    & \textbf{S}      & \textbf{S+R}     & \textbf{S}       & \textbf{S+R}       \\
\midrule
I2C                    & 89.2       & 90.0       & 88.5        & 88.3        & 87.1          & 87.5         \\
I2C2W                  & 94.3       & 94.5       & 95.2        & 95.1        & 95.0          & 94.8        \\
\bottomrule
\end{tabular}
\label{tab:ablation_normal}
\end{table}

The experiments in Tables \ref{tab:ablation_curved}-\ref{tab:ablation_normal} show that I2C (trained with two synthetic datasets) can achieve reasonable text recognition accuracy across nine public datasets. However, it captures little word semantic information and tends to fail when text images are degraded by poor lighting, low resolution, slanted views, etc., as illustrated in Fig. \ref{fig:sample}. C2W works from a different perspective which captures contextual linguistic information and complements I2C by identifying and correcting many false I2C detection effectively. The experiments in Tables 5-7 verify our ideas – I2C alone can achieve reasonable text recognition accuracy and including C2W further improves the accuracy by 4\% - 12\% across nine public datasets.

We also study how real images affect the scene text recognition. As Tables \ref{tab:ablation_curved}-\ref{tab:ablation_normal} shows, the fine-tuned model with real images improves the scene text recognition greatly for IC15, SVT, and SVTP whereas it does not work for the other 6 datasets. This is largely because images in IC15, SVT, and SVTP have clear blurs which are difficult to generate in synthetic images.



\subsection{Discussion}\label{sec:discussion}

In this section, we discuss the features of the proposed I2C2W including the effectiveness, significance, generalization ability, robustness, hyper-parameters, failure cases and efficiency.

\subsubsection{Effectiveness of I2C} 
The proposed I2C detects characters with their relative positions that are more significant to scene text recognition as compared with character localization with 2D coordinates. To demonstrate the effectiveness of I2C, we compare it with DETR \cite{carion2020end} on public scene text recognition datasets. Specifically, we train a character detector DETR on synthetic dataset which detects characters in text images with 2D coordinates. The detected characters are hence linked and sorted to words following \cite{liao2019scene}. As Table \ref{tab:detr} shows, the proposed I2C outperforms DETR by 10\%-15\% on different datasets, demonstrating the effectiveness of the proposed I2C. By further including the proposed C2W, the I2C2W outperforms DETR by large margins.

\begin{table}[!t]
\centering
\caption{\textbf{Comparison with DETR on scene text recognition}: The proposed I2C with relative position prediction clearly outperforms DETR on scene text recognition. By further including C2W, I2C2W achieves state-of-the-art performances.}
\begin{tabular}{cccccc}
\toprule
\textbf{Model} & \textbf{CUTE} & \textbf{TOTAL} & \textbf{SCUT-CTW} & \textbf{IC15} & \textbf{IC13} \\
\midrule
DETR           & 65.6  &  59.1   & 60.8   &  60.1    & 76.7          \\
I2C            & 81.2  &  71.0   & 72.4   &  75.2   &  87.1            \\
I2C2W          & 93.1  &  79.5   & 79.1   &  82.8   &  95.0           \\
\bottomrule
\end{tabular}
\label{tab:detr}
\end{table}

\subsubsection{Significance of C2W}
The proposed C2W complements I2C to detect characters more accurately. Fig. \ref{fig:sample} shows four sample images and their attention maps of different characters as produced by I2C2W, as well as the attention maps produced by I2C shown in the last column (only for falsely recognized characters by I2C). As Fig. \ref{fig:sample} shows, I2C fails to detect characters correctly when the characters suffer from background noises (e.g. `c' and `l' in the first two sample) or severe geometric distortions (e.g. `k' and `d' in the last two samples), largely because I2C predicts characters by using visual features of images only. C2W complements I2C to focus on not only each character but also their neighboring characters (i.e. contextual information), which helps to produce more accurate recognition as illustrated in Fig. \ref{fig:sample}.

\begin{table}[!t]
\centering
\caption{Scene text recognition performances over \textbf{license plate recognition dataset}: The proposed I2C2W outperforms the existing approaches clearly on contextless license plate recognition task. `Real': Real-world datasets.}
\begin{tabular}{llc}
\toprule
\textbf{Method}                                                     & \textbf{Data} & \textbf{Accuracy} \\ 
\midrule
SAR \cite{li2019show}                       & Synth90K+SynthText+Real            & 29.9              \\ 
ASTER \cite{shi2018aster}                   & Synth90K+SynthText            & 33.0              \\
SCATTER \cite{Litman_2020_CVPR}             & Synth90K+SynthText+SynthAdd            & 34.4              \\
AutoSTR \cite{zhang2020autostr}             & Synth90K+SynthText               & 35.1              \\
DAN \cite{wang2020decoupled}                & Synth90K+SynthText            & 51.1              \\ 
RobustScanner \cite{yue2020robustscanner}   & Synth90K+SynthText+Real            & 55.7              \\ 
\textbf{Ours}                               & Synth90K+SynthText            & \textbf{59.8}     \\ \bottomrule
\end{tabular}
\label{tab:comparison_car}
\end{table}

\begin{figure}[!t]
  \centering
  \includegraphics[width=\linewidth]{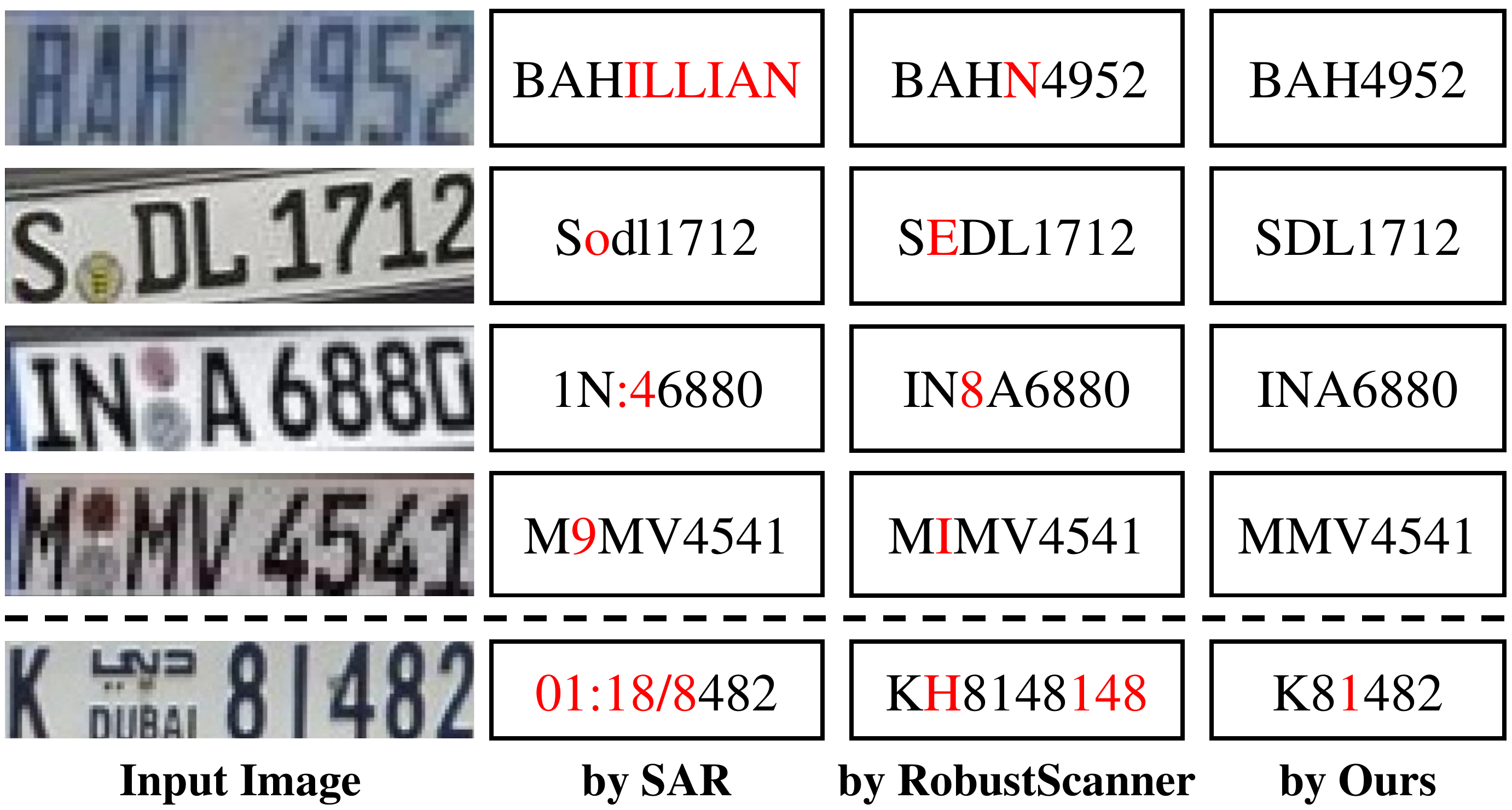}
\caption{
\textbf{Sample results from different approaches on license plate recognition dataset}: Given five input images, the SAR and RobustScanner usually produce incorrect or redundant characters in recognized words due to background noises and restriction of time steps. The proposed I2C2W is more tolerant to different noises and produces accurate results. The last row shows one typical failure case.
}
\label{fig:car}
\end{figure}

\subsubsection{Effectiveness of I2C2W}
We propose to detect characters through I2C in a non-sequential manner and then recognize words sequentially. To demonstrate the effectiveness of the proposed method, we conduct the experiments by training a I2W model (i.e. Backbone + Encoder + Decoder + CTC Head). Table \ref{tab:i2w} shows the results of I2C2W, I2C, I2W and two existing methods that adopt similar idea as I2W but have different implementations. The proposed I2C2W clearly outperforms all compared methods, demonstrating the effectiveness of the proposed two-stage recognition idea.


\begin{table}[!h]
\centering
\caption{\textbf{The effectiveness of I2C2W:} I2C2W clearly outperforms the one-stage method, demonstrating the its effectiveness.}
\begin{tabular}{cccc}
\toprule
\textbf{Model} & \textbf{IC15} & \textbf{SVTP} & \textbf{CUTE} \\
\midrule
Yang \textit{et at.} \cite{yang2020holistic} & 79.5          & 80.9          & 85.4          \\
Wan \textit{et al.} \cite{wan20192d}        & 76.3          & 82.3          & 86.8          \\ 
I2C           & 75.2         & 72.1          & 81.2  \\
I2W            & 79.8          & 81.7          & 87.8          \\
I2C2W          & 82.8          & 83.1          & 93.1      \\
\bottomrule
\end{tabular}
\label{tab:i2w}
\end{table}

\subsubsection{Generalization to contextless text recognition} 
The proposed C2W learns the character semantics which helps to improve the scene text recognition. This nice feature can be generalized to contextless text recognition task. To verify it, we follow \cite{yue2020robustscanner} to conduct an experiment on the Cars Dataset\cite{silva2018license} on license plate recognition task where texts on the license plate usually have much less semantic meaning. For a fair comparison with recent scene text recognition approaches, we use the model that is trained on synthetic data only without fine-tuning on any license plate images. As Table \ref{tab:comparison_car} shows, the proposed I2C2W outperforms existing approaches by 4.1\%, demonstrating the effectiveness of I2C2W on recognizing contextless data.

Additionally, Fig. \ref{fig:car} shows qualitative illustrations of SAR, RobustScanner and I2C2W on the Cars Dataset. As Fig. \ref{fig:car} shows, the proposed I2C2W produces more accurate recognition results for images with different background noises by training on Synthetic data only, demonstrating that I2C2W is more robust to recognize texts in images with different background. However, the I2C2W may fail when characters are very similar to each other (e.g. `i' and 'l' in the last row of Fig. \ref{fig:car}) or no contextual information can be used by C2W to correct falsely detected characters.

\begin{figure}[!t]
  \centering
  \includegraphics[width=\linewidth]{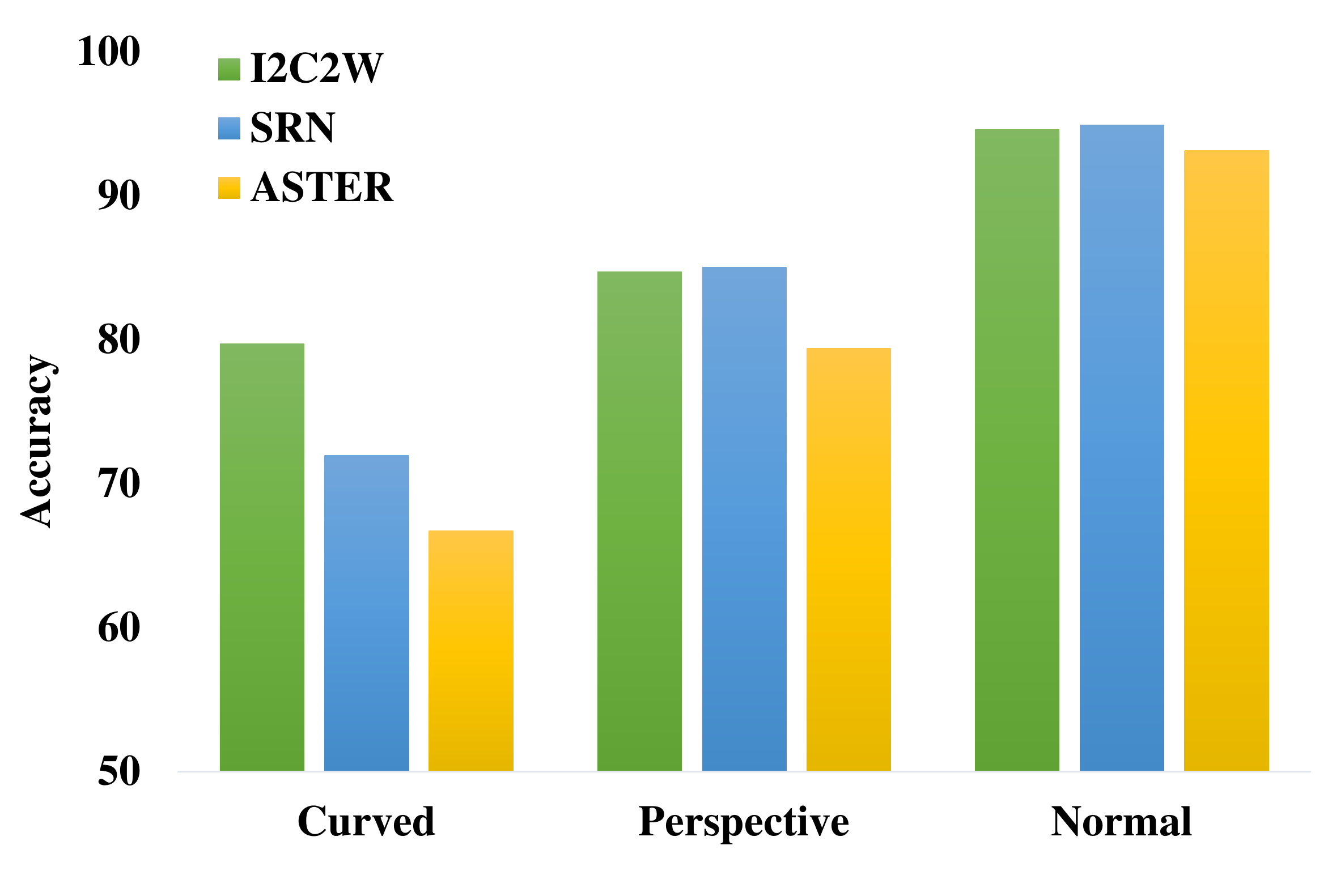}
\caption{\textbf{Mean accuracy on three types of datasets}: I2C2W obtains competitive accuracy on different types of datasets and it clearly outperforms the state-of-the-art on curved datasets.}
\label{fig:mean}
\end{figure}

\subsubsection{Robustness of I2C2W}
To demonstrate the robustness of I2C2W, we show the mean accuracy of I2C2W, SRN, and ASTER on all nine public datasets that are trained under the same settings. As Fig. \ref{fig:mean} shows, I2C2W achieves competitive accuracy over all three types of datasets, demonstrating the robustness of I2C2W on recognizing curved, multi-oriented, and normal texts in scene images. Additionally, I2C2W outperforms existing work by large margins on curved texts. This shows that I2C2W is much more robust to recognize texts with large shape variations as compared with state-of-the-art methods.

\subsubsection{Hyper-Parameter $N$}
We study the influence of $N$ on the Total-Text dataset where $N$ refers to the total number of detected character candidates by I2C. As Table \ref{tab:hyper} shows, I2C2W achieves the best accuracy when $N$ is 25. Reducing $N$ to 10 lowers the accuracy significantly as many words contains more than 10 characters which directly translates to many miss detection. At the other end, increasing $N$ to 50 or 100 degraded the scene text recognition in most cases, largely because the additionally included noisy character candidates (by a $N$ larger than 25) introduce more confusion while training C2W.


\begin{table}[!t]
\centering
\caption{The accuracy of I2C2W on Total-Text dataset \textbf{under different hyper-parameter $N$}.}
\begin{tabular}{ccccc}
\toprule
\multirow{2}{*}{\textbf{Dataset}} & \multicolumn{4}{c}{\textbf{Hyper-parameter $N$}} \\ \cmidrule(lr){2-5}
                                & \textbf{10} & \textbf{25} & \textbf{50} & \textbf{100}\\
\midrule
\textbf{CUTE}                & 89.6  & \textbf{93.1}  & \textbf{93.1}  & 89.6 \\
\textbf{TOTAL}               & 76.7  & \textbf{79.8}  & 79.1  & 79.0 \\
\textbf{SCUT-CTW}            & 73.6  & \textbf{79.1}  & 78.3  & 78.5 \\
\textbf{IC15}                & 74.9  & \textbf{82.8}  & 81.5  & 81.0\\
\textbf{SVT}                 & 84.4  & \textbf{91.7}  & 89.5  & 89.5\\
\textbf{SVTP}                & 74.0  & \textbf{83.1}  & \textbf{83.1}  & 81.4\\
\textbf{IIIT5K}              & 89.5  & \textbf{94.3}  & 93.1  & 92.7\\
\textbf{IC03}                & 90.1  & \textbf{95.2}  & 93.3  & 93.3 \\
\textbf{IC13}                & 86.7  & \textbf{95.0}  & 93.5  & 92.8\\
\bottomrule
\end{tabular}
\label{tab:hyper}
\end{table}

\subsubsection{Failure Cases} 

\begin{figure}[!h]
  \centering
  \includegraphics[width=\linewidth]{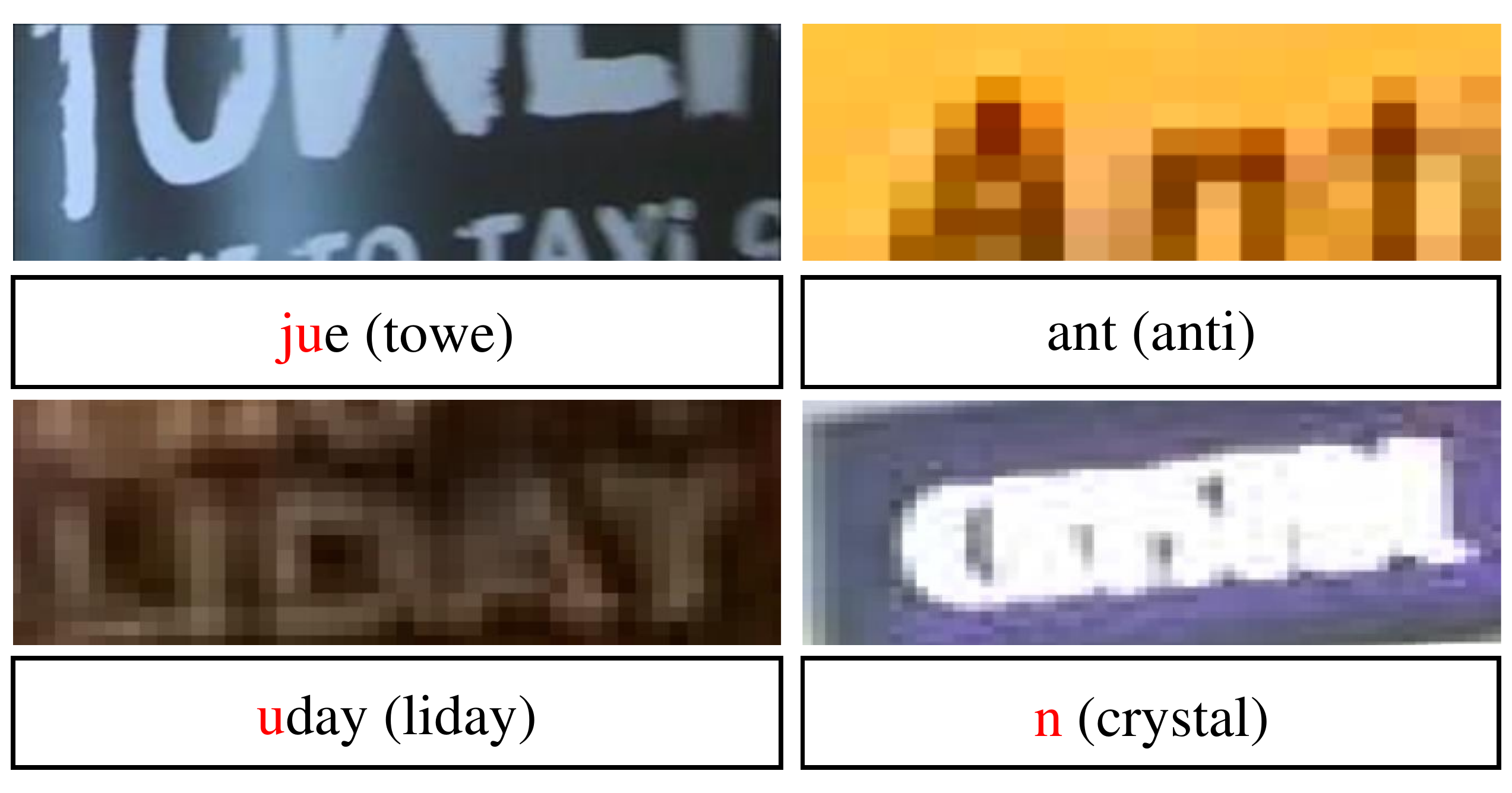}
\caption{
\textbf{Illustration of typical failure cases}: Given four sample images, the recognition results from I2C2W and corresponding ground-truth are shown under each input image. The wrong recognized characters are highlighted in red. I2C2W tends to fail under the presence of heavy occlusions, ultra-low resolution, etc. 
}
\label{fig:fail}
\end{figure}

The proposed I2C2W usually fails under several typical scenarios as shown in Fig. \ref{fig:fail}. First, the proposed I2C2W may fail if the scene texts suffer from heavy occlusions. Only part of texts is shown in the image which usually leads to wrong recognition of characters as illustrated in the first row of Fig. \ref{fig:fail}. Second, it may produce incorrect recognition if the scene texts suffer from strong blurs or ultra-low resolution where the characters are difficult to be separated from each other as shown in the second row of Fig. \ref{fig:fail}.

\subsubsection{Efficiency} 
I2C2W can process an image with around 31ms on average for the dataset ICDAR2013. The processing speed is comparable with SRN’s 30ms per image and slightly slower than ASTER’s 20ms per image with the same workstation with a single NVIDIA Telsa V100. By adopting a parallel attention mechanism, I2C2W could be faster than ASTER while handling long text lines as studied in \cite{yu2020towards}.


\section{Conclusion and Future Work}\label{sec:conclusion}
This paper presents an accurate and robust text recognizer I2C2W. It introduces an image-to-character module (I2C) that detects characters and predicts their relative positions in words non-sequentially. By ignoring the restrictions of time steps, I2C strives to detect all possible characters with different features alignments. It also designs a character-to-word module (C2W) that learns character semantics from the detected characters by I2C to produce word recognition. The proposed I2C and C2W complement each other and can be trained end-to-end. Extensive experiments over nine public datasets show that I2C2W achieves state-of-the-art performances over normal datasets and outperforms existing approaches by a large margin on curved datasets, demonstrating the effectiveness of the proposed image-to-character-to-word recognition approach.

In the future, we would like to continue to study scene text recognition by learning character and word semantics. Specifically, we plan to leverage the advances in NLP tasks like character or word embeddings that model the semantics of texts well. We also would like to study the feasibility of these NLP advances on complementing visual feature extraction for accurate scene text recognition.

\bibliographystyle{IEEEtran}
\bibliography{IEEEtran}
\end{document}